\documentclass{article}

\usepackage[final,nonatbib]{neurips_2019}

\usepackage[utf8]{inputenc} 
\usepackage[T1]{fontenc}    
\usepackage{hyperref}       
\usepackage{url}            
\usepackage{booktabs}       
\usepackage{amsfonts}       
\usepackage{nicefrac}       
\usepackage{microtype}      
\usepackage{graphicx}
\usepackage{subfigure}
\usepackage{cite}
\usepackage{algorithm}
\usepackage{algorithmic}
\usepackage{amsmath}

\title{Ranking architectures using meta-learning}

\author{
\hskip-0.6in Alina Dubatovka \\
\hskip-0.6in ETH Zürich \\
\hskip-0.6in \texttt{alina.dubatovka@inf.ethz.ch\thanks{Work done during an internship at Google.}} \\
\And
\hskip-0.5in Efi Kokiopoulou  \\
\hskip-0.5in Google AI \\
\hskip-0.5in \texttt{efi@google.com} \\
\AND
Luciano Sbaiz \\
Google AI \\
\texttt{sbaiz@google.com} \\
\And
Andrea Gesmundo \\
Google AI \\
\texttt{agesmundo@google.com} \\
\AND
G\'abor Bart\'ok \\
Google AI \\
\texttt{bartok@google.com} \\
\And
Jesse Berent \\
Google AI \\
\texttt{jberent@google.com} \\
}

\begin{document}

\maketitle              
\begin{abstract}
Neural architecture search has recently attracted lots of research efforts as it promises to automate the manual design of neural networks. However, it requires a large amount of computing resources and in order to alleviate this, a performance prediction network has been recently proposed that enables efficient architecture search by forecasting the performance of candidate architectures, instead of relying on actual model training. The performance predictor is task-aware taking as input not only the candidate architecture but also task meta-features and it has been designed to collectively learn from several  tasks. In this work, we introduce a pairwise ranking loss for training a network able to rank candidate architectures for a new unseen task conditioning on its task meta-features. We present experimental results, showing that the ranking network is more effective in architecture search than the previously proposed performance predictor.

\end{abstract}






\section{Introduction }

The design of neural networks specific to a certain task requires a considerable amount of work. To alleviate this, a number of solutions have been proposed for neural architecture search~\cite{ZophLe16,MeKlFeuSpHut16,ReMoSeSaSuTaLeKu17,DARTS,ENAS,BeKiZoVaLe18,GraphHyperNets19,ProxyLessNAS}. Despite their effectiveness, these solutions require to explore a large number of architectures in order to maximize performance. To reduce the computational cost, a performance prediction network has been introduced for fast architecture search~\cite{KoHaSbGeBaBe19}. The network predicts the candidate network performance and allows performing architecture search with a simple gradient-ascent optimization.

In this work we introduce a network able to rank architectures. This is used instead of the performance prediction network in the gradient-ascent procedure. The network takes as input a representation of the architecture and a set of meta-features derived from the task and produces a score for the architecture. The ranking loss enforces that good architectures produce higher scores than poor architectures. Similarly to~\cite{KoHaSbGeBaBe19}, the ranking network is co-trained with the sub-network that generates the task meta-features.
We present experimental results, showing that the ranking network significantly outperforms the performance prediction network in terms of test accuracy of the final found architecture.

\section{Background}
\label{sec:framework}

\begin{figure}[t]
    \centering
        \includegraphics[scale=0.43]{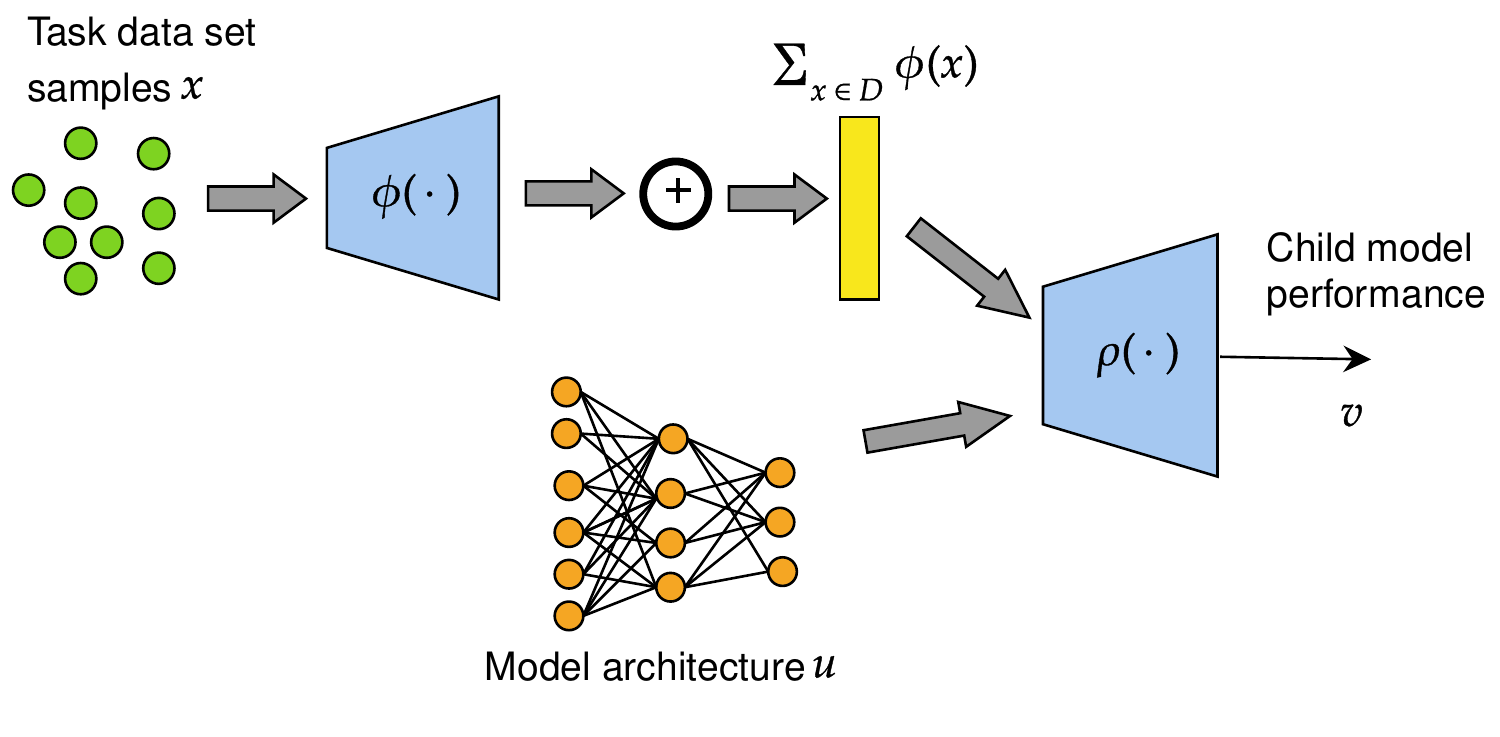}
    \caption
    {The architecture of the performance predictor, which consists of $\phi(\cdot)$ and $\rho(\cdot)$.
    }
    \label{fig:dvn}
\end{figure}

We start by reviewing the previously proposed framework in \cite{KoHaSbGeBaBe19}. The framework lies on a performance prediction network that estimates the performance of a candidate architecture on a certain task.
On a high level, the performance predictor acts as a meta-model that helps in tuning the architecture of a child model. In what follows, we consider child model families parametrized by $u$, assuming for now that $u$ is a vector of continuous variables.
The performance prediction network takes as input: (i) descriptive meta-features $z$ derived from a certain task data set and (ii) the child model architecture parameters $u$,
and predicts how well the architecture $u$ performs on the task data set described by $z$. The performance metric $v$ may take various forms (e.g., accuracy, AUC). The network is shown conceptually in Fig. \ref{fig:dvn}.

The meta-features are learned directly from the task data set $D$. The data set (or a large fraction of it) is given as input to the performance predictor, and a task embedding is learned directly from the raw task data set samples. 
This task embedding plays the role of the meta-features and is learned jointly together with the rest of the weights of the prediction network.
In order to ensure that the order of the samples does not matter, the task embedding is decomposed in the form $\sum_{x \in D} \phi(x)$ for suitable transformation $\phi$, where the latter is typically implemented by a few layers (e.g., fully connected, non-linearities etc.).
Each sample from the task data set is transformed  using $\phi(\cdot)$ and then aggregated into the task embedding.
This process is shown conceptually in Fig.~\ref{fig:dvn}, where the performance prediction network essentially consists of $\phi(\cdot)$ and $\rho(\cdot)$ that are jointly learned, \textit{i.e.},
$v(u, z) := \rho\left(u, \sum_{x \in D} \phi(x)\right)$.

Assuming there are $K$ tasks, the framework in its offline phase generates, for each task, several child model architectures, trains them and collects the model performances on the validation set in a database of model training experiments.
This database is used to generate the training set for the performance prediction network.
Once the performance prediction network $v(u, z; w)$ has been trained, the model weights $w$ are kept fixed. In the online phase, given a new unseen task dataset with meta-features $z$,
the framework utilizes the gradient of $v(u, z)$ with respect to $u$ in order to perform gradient-based optimization and get a good candidate architecture $\hat{u}$ that maximizes the estimated child model performance.

Note also that in order to be able to perform gradient-based inference, the model architecture parameters $u$ need to live in a continuous space. Previous work achieves that by moving away from the categorical nature of design choices using a parametrized softmax over all possible choices.
For example, given a basis set consisting of $n$ base layers $o_i(x)$ corresponding to different sizes and activation functions, one can associate a weight $\alpha_i$ with each base layer, and define a new parametrized layer: 
 $o(x) = \sum_{i=1}^{n} \frac{\exp(\alpha_i)}{\sum_{j=1}^{n} \exp(\alpha_j)} o_i(x).$
The same softmax trick can be applied for combining several such parametrized layers together or for combining several embedding modules together. Please see \cite{KoHaSbGeBaBe19} for more details about the continuous parametrization of the child networks.

\section{Ranking loss}
\label{sec:ranking_loss}



The goal of architecture search is to find the model architecture with the best performance.
Predicting the actual value of the child model performances is not really required; what is actually needed is a way to rank different candidate architectures. Therefore, a ranking loss seems more suitable for this problem instead of relying on traditional L1 or L2 losses. Motivated by this observation, we introduce a pairwise ranking loss that encourages top architectures score higher than poorly performing architectures. We introduce first two important ingredients of our loss definition:

\begin{itemize}
    \item \textbf{Margin}. In order to encourage the network to produce more distinguishable rankings, we introduce a margin term $m$ which specifies how much the ranking values should differ in order to be called sufficiently different. 
\item \textbf{Uncertainty gap}. The ground truth values represent performances of different child model architectures on the validation set of a certain task. These values are typically noisy measurements due to the training stochasticity. Taking into consideration this uncertainty, we call some pairs “indistinguishable” when their performance values are within the uncertainty gap.
Such pairs do not contribute to the final loss. 

\end{itemize}

\paragraph{Pairwise ranking loss}
Putting margin and uncertainty gap parameters together, we get the following formula for the pairwise ranking loss:
\[\label{eq:linear_ranking_loss}
  L(v_i, v_j) = \left\{
  \begin{array}{ll}
    \max(0, ~m - (v_i - v_j)),  &\textrm{if ~}  p_i - p_j > \textrm{gap} \\
    0, &  \textrm{otherwise}
  \end{array}\right.
\]

where $p_i$, $p_j$ is the ground truth performances of architectures $i$ and $j$ respectively. We call the loss above \textbf{linear ranking loss} since the scores are linearly related to the loss value.


One issue with the linear ranking loss is that its gradient is constant, implying that we treat small discrepancies
the same as high discrepancies, which may not be ideal. Therefore, we also propose a \textbf{quadratic ranking loss} function, defined as:
\[\label{eq:quadratic_ranking_loss}
  Q(v_i, v_j) = \left\{
  \begin{array}{ll}
\frac{\max(0, ~m - (v_i -v_j))^2}{m} ,  &\textrm{if ~}  p_i - p_j > \textrm{gap} \\
    0, &  \textrm{otherwise}
  \end{array}\right.
\]
The reason why we divide by the margin $m$ is because we don't want the loss to scale quadratically with the margin value.

\section{Experiments}
\label{sec:experiments}


\paragraph{\bf Setup.}
We follow the setup in~\cite{KoHaSbGeBaBe19} using publicly available NLP data sets, which is shortly summarized below.
The child models have been implemented using the parametrization described in Sec.~\ref{sec:framework}. 
The sizes of the base layers in a single parametrized layer are $\{8, 16, 32, 64, 128, 256\}$ and each one of them is combined with two distinct activation functions (\texttt{relu} and \texttt{tanh}). Hence a single parametrized layer is composed of twelve base layers and each child model has seven such parametrized layers.
In addition, each child model has seven text embedding modules. Hence, the resulting architecture search space consists of $7 + 7 \cdot 12 + 2 \cdot 7 = 105$  dimensions, which is rather high-dimensional.

The performance prediction network was trained on the child model training experiments stored in the database, which was populated with about 500 random child model architectures per task.
We used a small network consisting of two fully connected layers of size 50 each for the task meta-features tower (aka $\phi(\cdot)$ in Fig. \ref{fig:dvn}) and two fully connected layers of sizes 50 and 10 for the tower that produces the final prediction (aka $\rho(\cdot)$ in Fig.~\ref{fig:dvn}).
In contrast to the original setup, we trained the network with the ranking losses proposed above (both linear and quadratic versions). For all the experiments, we set uncertainty gap to 0.01, since it roughly corresponds to the noise level of measurements in our database (please see the Appendix for more details). This prevents unreliable pairs from contributing to the ranking loss. The margin was set to 0.3 according to cross-validation experiments. 
We did not perform any further hyper parameter search. The ranking loss is optimized using Stochastic Gradient Descent with momentum~\cite{momentum} (using 0.5 as default parameter). The learning rate was set to $10^{-4}$ according to common practices.

\subsection{Predicting the model architecture performance}

\begin{table*}[t]
\begin{center}
\begin{small}
\begin{sc}
\begin{tabular}{lccc}
\toprule
Task name & L2-Loss & Linear Ranking Loss & Quadratic Ranking Loss \\
 \midrule
airline & 0.7454 $\pm$ 0.0237 & 0.9042 $\pm$ 0.0134 & \textbf{0.9152 $\pm$ 0.0238} \\
emotion & 0.7126 $\pm$ 0.0189 & \textbf{0.9023 $\pm$ 0.0053} & 0.8918 $\pm$ 0.0298 \\
global warming & 0.6633 $\pm$ 0.0204 & \textbf{0.8701 $\pm$ 0.0057} & 0.8591 $\pm$ 0.0190 \\
corporate messaging & 0.6316 $\pm$ 0.0157 & \textbf{0.9148 $\pm$ 0.0056} & \textbf{0.9190 $\pm$ 0.0076} \\
disasters & 0.6613 $\pm$ 0.0272 & 0.8677 $\pm$ 0.0158 & \textbf{0.8880 $\pm$ 0.0146} \\
political message & 0.3403 $\pm$ 0.0178 & 0.6354 $\pm$ 0.0296 & \textbf{0.6981 $\pm$ 0.0305} \\
political bias & 0.3643 $\pm$ 0.0100 & 0.5770 $\pm$ 0.0089 & \textbf{0.6069 $\pm$ 0.0227} \\
progressive opinion & 0.6626 $\pm$ 0.0239 & \textbf{0.8454 $\pm$ 0.0052} & 0.8330 $\pm$ 0.0122 \\
progressive stance & 0.5969 $\pm$ 0.0122 & \textbf{0.8628 $\pm$ 0.0123} & 0.8413 $\pm$ 0.0110 \\
us economy & 0.1817 $\pm$ 0.0091 & 0.4198 $\pm$ 0.0361 & \textbf{0.5902 $\pm$ 0.0739} \\
\bottomrule
\end{tabular}
\end{sc}
\end{small}
\end{center}
\vskip-0.1in
\caption{Behaviour of the Spearman's rank correlations values; breakdown by task.\label{table:Spearman_values}}
\end{table*}

We have performed several leave-one-out experiments, where each task in our set is considered to be a test task and  the rest of the tasks being used as the training tasks. Then for each such leave-one-out experiment, we train a ranking network and study its predictive performance.
In particular, given the predicted performances and their corresponding actual performances, we quantify the predictive  performance in terms of the Spearman's rank correlation coefficient.
Table~\ref{table:Spearman_values} shows a comparison of L2, linear and quadratic ranking loss functions according to  Spearman's rank correlations for each task.
Notice that introducing the ranking loss significantly improves the predictive performance of the network.


\subsection{Architecture search}

\begin{table*}[t]
\begin{center}
\begin{small}
\begin{sc}
\begin{tabular}{lcccc}
\toprule
Task name & NAS & L2-Loss & Linear & Quadratic \\
\midrule
airline & 0.83197 & 0.8260 $\pm$ 0.0043 & 0.8227 $\pm$ 0.0000 & \textbf{0.8278 $\pm$ 0.0000} \\
global warming & 0.79196 & 0.8066 $\pm$ 0.0159 & \textbf{0.8182 $\pm$ 0.0001} & 0.8121 $\pm$ 0.0002 \\
disasters & 0.83425 & 0.8242 $\pm$ 0.0082 & \textbf{0.8317 $\pm$ 0.0000} & \textbf{0.8313 $\pm$ 0.0001} \\
political bias & 0.778 & 0.7728 $\pm$ 0.0161 & \textbf{0.7814 $\pm$ 0.0001} & 0.7710 $\pm$ 0.0000 \\
progressive opinion & 0.73276 & 0.7250 $\pm$ 0.0191 & \textbf{0.7345 $\pm$ 0.0001} & 0.7124 $\pm$ 0.0002\\
progressive stance & 0.57759 & 0.5162 $\pm$ 0.0222 & 0.4491 $\pm$ 0.0020 & \textbf{0.5230 $\pm$ 0.0354} \\
us economy & 0.76411 & 0.7509 $\pm$ 0.0140 & \textbf{0.7688 $\pm$ 0.0001} & 0.7540 $\pm$ 0.0002 \\
corporate messaging & 0.85897 & 0.8519 $\pm$ 0.0262 & \textbf{0.8726 $\pm$ 0.0001} & 0.8708 $\pm$ 0.0002 \\
emotion &  0.35425 & \textbf{0.3480 $\pm$ 0.0238} & 0.3256 $\pm$ 0.0000 & 0.3394 $\pm$ 0.0007 \\
political message & 0.414 & 0.4264 $\pm$ 0.0070 & \textbf{0.4281 $\pm$ 0.0000} & 0.4231 $\pm$ 0.0000 \\
\bottomrule
\end{tabular}
\end{sc}
\end{small}
\end{center}
\vskip -0.1in
\caption{The test accuracy achieved by the found architectures. We illustrate in boldface the best performance among the three loss functions we experimented with.}
  \label{table:inference_results}
\end{table*}

We also looked into the quality of the child model architectures discovered by the ranking network in terms of test accuracy. 
When we apply our gradient ascent optimization we pick the initial guesses using the top five architectures from the two closest training tasks in the task embedding space. 
Each experiment is repeated ten times (including training the prediction network from scratch ten times) in order to get more accurate statistics on the performances. 
Table \ref{table:inference_results} shows the architecture search results in terms of test accuracy. Notice that the ranking network outperforms the L2 performance predictor in the majority of the cases.
We also include a comparison with the NAS method \cite{ZophLe16} in order to get our results in perspective.
In this case, NAS is applied for each test task independently and it typically requires a large number (higher than 1000) of child model trainings  in order to optimize the validation performance. 
On the other hand, our method is able to identify a good architecture \textit{before} any child model training is performed.



\section{Conclusions and future work}
\label{sec:conclusion}
We introduced a ranking network that is able to effectively rank candidate architectures and propose good architectures for new tasks.
The proposed approach remains fast thanks to the efficient gradient ascent applied in the architecture space, which does not require any child model trainings or any intermediate child model weight updates.
In our future work, we plan to explore efficient solutions for populating the database of model training experiments.


\bibliographystyle{plain}
\bibliography{references}

\newpage
\appendix
\section{Appendix}

\vskip 0.3in
\subsection{Datasets}

Table~\ref{NLP-datasets-table} summarizes main characteristics of the NLP datasets used for the experimental training and validation.

\begin{table*}[htbp]
\begin{center}
\begin{small}
\begin{tabular}{lccccc}
\toprule
\sc{Data set} & \sc{Train Examples} & \sc{Val. Examples} & \sc{Test Examples} & \sc{Classes} \\
\midrule
\sc{Airline}                 & 11712 & 1464 & 1464 & 3   \\
\sc{Corporate Messaging}     & 2494  & 312 & 312   & 4   \\
\sc{Emotion}                 & 32000 & 4000 & 4000 & 13  \\
\sc{Disasters}               & 8688  & 1086 & 1086 & 2   \\
\sc{Global Warming}          & 3380  & 422  & 423  & 2   \\
\sc{Political Bias}          & 4000  & 500  & 500  & 2   \\
\sc{Political Message}       & 4000  & 500  & 500  & 9   \\
\sc{Progressive Opinion}     & 927   & 116  & 116  & 3   \\
\sc{Progressive Stance}      & 927   & 116  & 116  & 4   \\
\sc{US Economy} & 3961  & 495  & 496  & 2   \\
\bottomrule
\end{tabular}
\end{small}
\end{center}
\caption{Statistics for the NLP classification data sets. Number of examples in the training set, validation set and test set and number of classes. All data sets are publicly available from \texttt{crowdflower.com}. \label{NLP-datasets-table}}
\end{table*}

\vskip 0.3in
\subsection{Impact of ranking loss to Pearson correlation values}

We have seen that when we switch from L2 loss to the ranking loss, the Spearman's rank correlation values between the actual performances and the predicted performances increases. For the sake of completeness, Table \ref{table:Pearson_values} shows the predictive performance of the prediction network measured by Pearson correlation as well. The results confirm that the ranking loss helps the  network to make better predictions as both metrics (Spearman and Pearson) are improving.

\begin{table*}[htbp]
\begin{center}
\begin{small}
\begin{sc}
\begin{tabular}{lcccc}
\toprule
Task name & L2-Loss & Linear Ranking Loss & Quadratic Ranking Loss \\
\midrule
airline & 0.9037 $\pm$  0.0108 & 0.9627 $\pm$ 0.0054 & \textbf{0.9848 $\pm$ 0.0134} \\
emotion & 0.8230 $\pm$ 0.0077 & 0.9524 $\pm$ 0.0049 & \textbf{0.9599 $\pm$ 0.0078} \\
global warming & 0.7539 $\pm$ 0.0100 & \textbf{0.9303 $\pm$ 0.0031} & 0.9225 $\pm$ 0.0181 \\
corporate messaging & 0.6299 $\pm$ 0.0101 & 0.9378 $\pm$ 0.0099 & \textbf{0.9441 $\pm$ 0.0059} \\
disasters & 0.8443 $\pm$ 0.0168 & 0.9327 $\pm$ 0.0285 & \textbf{0.9881 $\pm$ 0.0030} \\
political message & 0.7820 $\pm$ 0.0211 & 0.9190 $\pm$ 0.0247 & \textbf{0.9844 $\pm$ 0.0042} \\
political bias & 0.3338 $\pm$ 0.0105 & \textbf{0.5385 $\pm$ 0.0086} & 0.5332 $\pm$ 0.0135 \\
progressive opinion & 0.5588 $\pm$ 0.0119 & \textbf{0.8455 $\pm$ 0.0031} & 0.8086 $\pm$ 0.0311 \\
progressive stance & 0.5276 $\pm$ 0.0137  & \textbf{0.8460 $\pm$ 0.0153} & 0.8113 $\pm$ 0.0109 \\
us economy & 0.7181 $\pm$ 0.0238 & 0.8821 $\pm$ 0.0455 & \textbf{0.9623 $\pm$ 0.0124} \\
\bottomrule
\end{tabular}
\end{sc}
\end{small}
\end{center}
\caption{Pearson correlation values between the actual performances and the scores provided by the ranker; breakdown by task. The higher the better.}
  \label{table:Pearson_values}
\end{table*}

\vskip 0.3in
\subsection{Uncertainty gap}

\begin{figure*}[htbp]
\centering
\includegraphics[scale=.4]{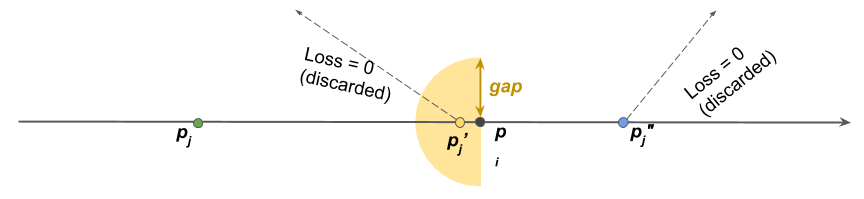}
\caption{Effect of the uncertainty gap value on the loss.}
\label{fig:uncertainty_gap}
\end{figure*}

Recall that the uncertainty gap accounts for noise in measuring the performance of an architecture. When the performances of two architectures is within the uncertainty gap we consider them as "indistinguishable" and we exclude them from the final loss. This is shown in Fig.  \ref{fig:uncertainty_gap}. Notice that only the pair $(p_i, p_j)$ contributes to the ranking loss. The pair $(p_i, p_j'')$ is excluded, since we only count pairs with $p_i <p_j$ to ensure each pair is counted once. The pair $(p_i, p_j')$ is also excluded, since the difference between $p_i$ and $p_j'$ is less than the uncertainty gap.

\begin{figure*}[htbp]
\centering
\includegraphics[scale=.4]{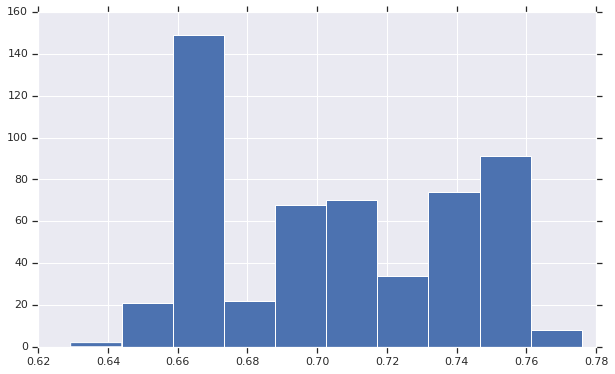}
\caption{Distribution of the child model performances for the \textsc{progressive opinion} task.}
\label{fig:gap}
\end{figure*}

Figure \ref{fig:gap} shows the distribution of child model performances for one of the tasks. One can see that many samples lie within 0.01 interval. Based on this empirical evidence, we chose the uncertainty gap to be 0.01 in our experiments.

\vskip 0.3in
\subsection{Discussion: linear vs quadratic ranking loss}

\begin{figure*}[htbp]
\centering
\includegraphics[scale=.5]{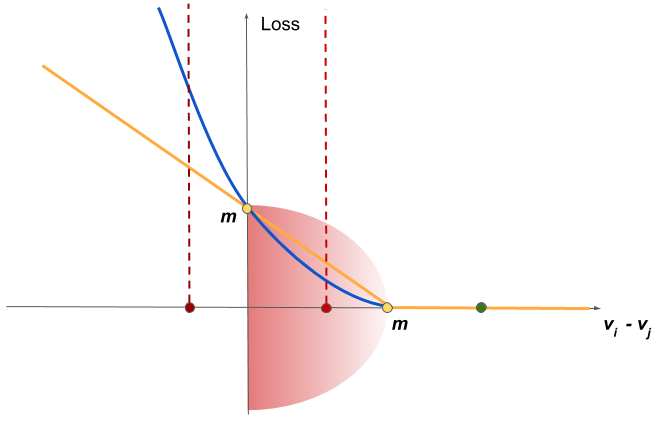}
\caption{Behaviour of the Linear (yellow line) and Quadratic (blue line) ranking loss functions against the score difference $v_i - v_j$.}
\label{fig:losses}
\end{figure*}

Despite the fact that the quadratic function may be a more intuitive choice for the loss function, as discussed in the main text, the experimental results show that this may not always be the case.
One possible explanation might be the fact that the quadratic loss is sensitive to the combination of architecture performances included in the batch. 
For those pairs that passed the uncertainty gap filtering, the gradient of the quadratic loss depends on their pairwise differences, i.e.,
$$\frac{\partial Q}{\partial v_i} = \sum_{v_j > v_i - m} \frac{2}{m} \cdot (m - (v_i - v_j)) \cdot -1 $$
$$ \frac{\partial Q}{\partial v_j} = \sum_{v_i < v_j + m} \frac{2}{m} \cdot (m - (v_i - v_j)) \cdot 1 $$
On the other hand, for those pairs, the gradient of the linear loss is constant, i.e.,
$$\frac{\partial L}{\partial v_i} = \sum_{v_j > v_i - m} -1 $$
$$\frac{\partial L}{\partial v_j} = \sum_{v_i < v_j + m} 1 $$ 

This very fact implies that the gradients of the quadratic loss may be of high variance, which may result in training instability and optimization difficulties.


\vskip 0.3in
\subsection{Visualization of the task meta-features}

We looked into the learned task representations in the meta-feature space. In particular, for each task we computed
the task embedding for different batch realizations (of the task samples) and we visualized them both with tSNE as well as with Principal Component Analysis (PCA).

\begin{figure*}[htbp]
\centering
\includegraphics[scale=.4]{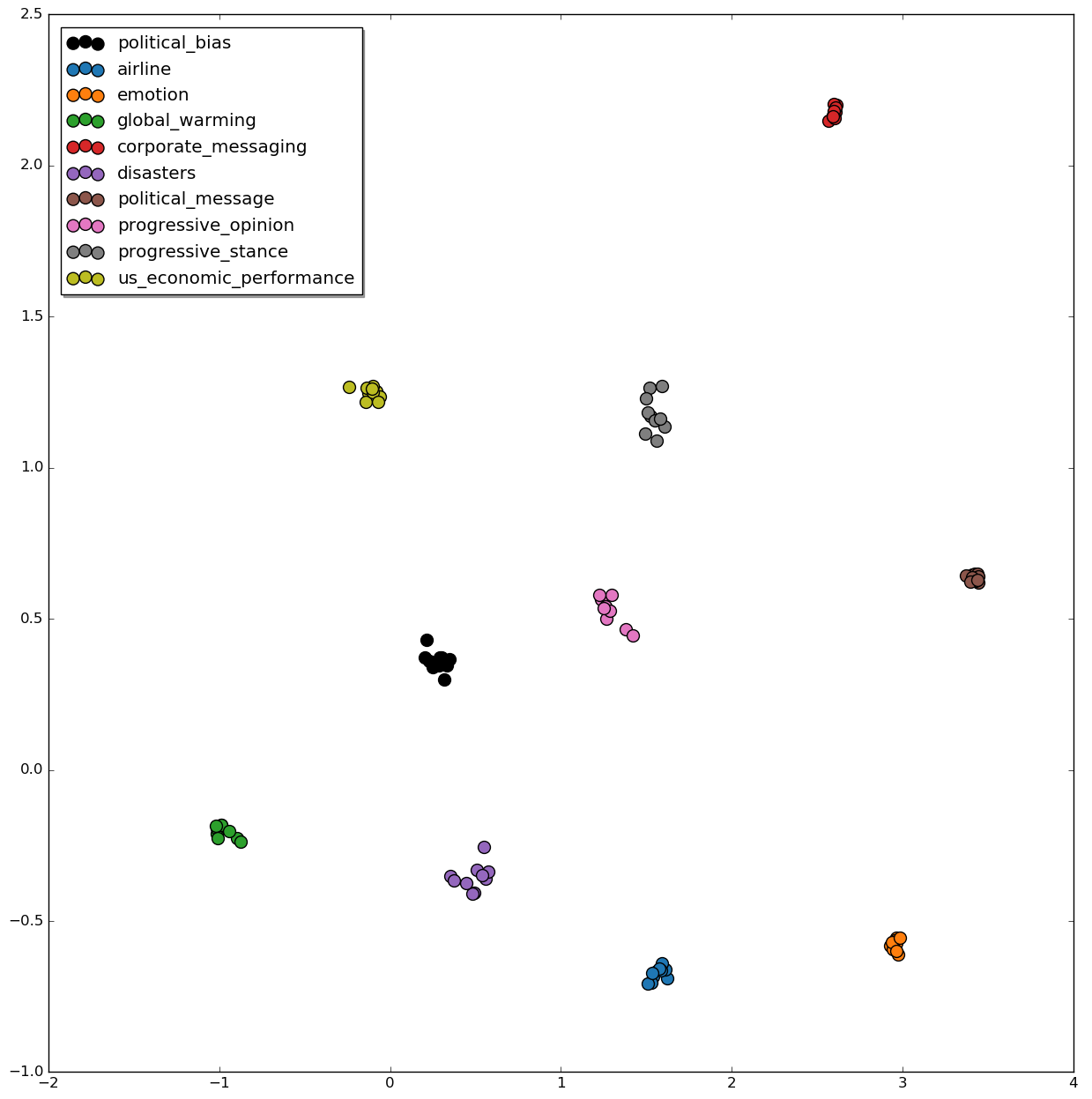}
\caption{Visualization of the learned meta-features using tSNE for the test task \textsc{political bias} (black color) and the training tasks (other colors). For each task we show the meta-features computed from 10 random batches of the task samples.}
\label{fig:tSNE}
\end{figure*}

\begin{figure*}[htbp]
\centering
\includegraphics[scale=.4]{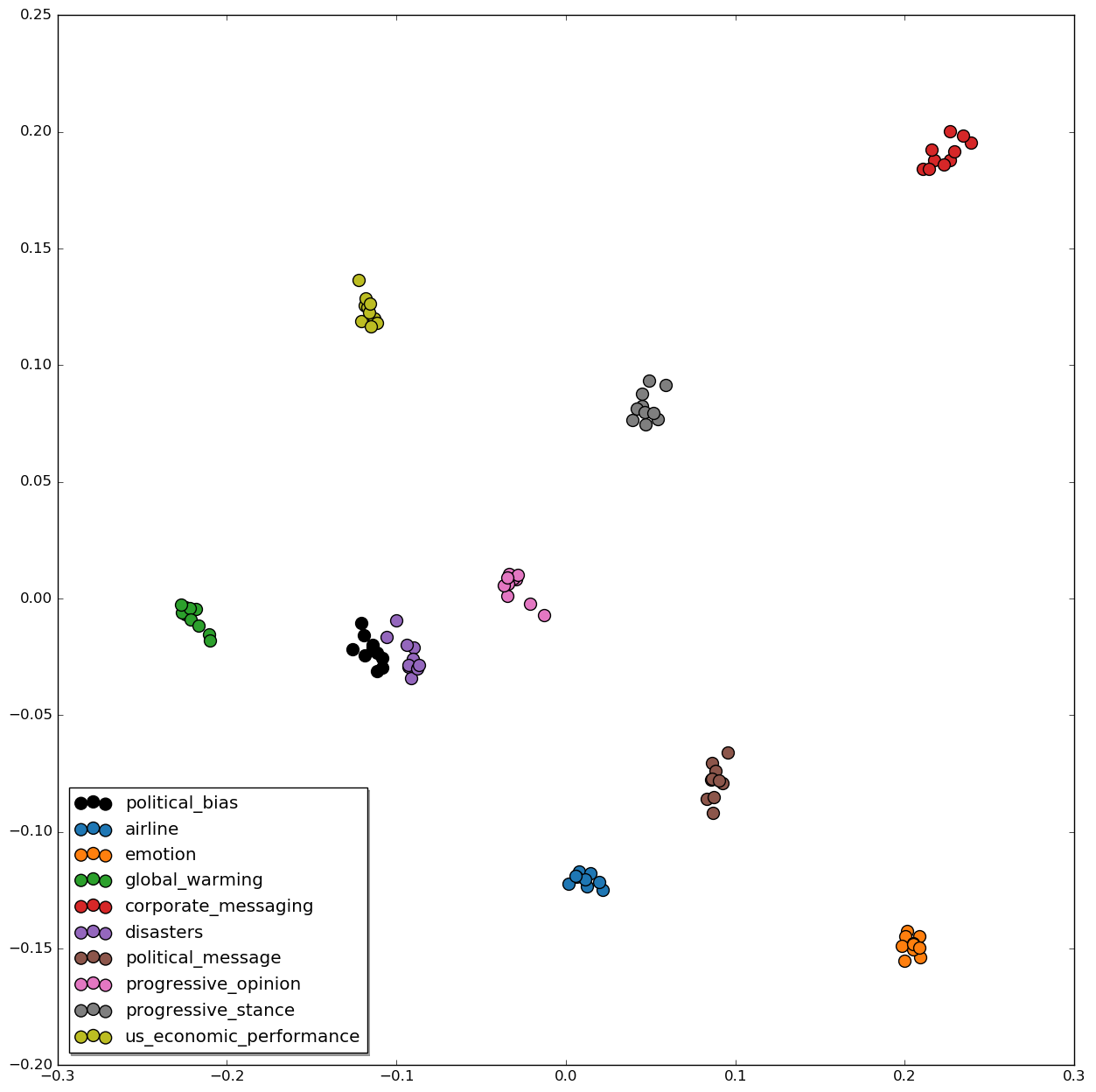}
\caption{Visualization of the learned meta-features using PCA for the same setup as in Fig.~\ref{fig:tSNE}.}
\label{fig:PCA}
\end{figure*}

Figures \ref{fig:tSNE} and \ref{fig:PCA} show the 2D visualizations for tSNE and PCA respectively for 10 random batches of the training and the test tasks. For tSNE we set the perplexity to 70.
Interestingly, different batch realizations from the same task result in close-by task embeddings in the meta-feature space, which confirms the stability of the method in this respect.

\vskip 0.3in
\subsection{Text input embedding modules}

Table \ref{table:hub_modules} shows the Tensorflow Hub embedding modules that we used for text input in our experiments.

\begin{table*}[htbp]
\vskip 0.15in
\begin{center}
\begin{small}
\begin{tabular}{lccl}
\toprule
Language & Dataset size & Embed dim.  & TensorFlow Hub Handles  \\
         &              &             & Prefix: \texttt{https://tfhub.dev/google/...}  \\
\midrule
English & 4B   & 250 & \href{https://tfhub.dev/google/Wiki-words-250/1}{\texttt{Wiki-words-250/1}}  \\
English & 200B & 128 & \href{https://tfhub.dev/google/nnlm-en-dim128/1}{\texttt{nnlm-en-dim128/1}}  \\
English & 7B   & 50  & \href{https://tfhub.dev/google/nnlm-en-dim50/1}{\texttt{nnlm-en-dim50/1}} \\
English & -    & 512 & \href{https://tfhub.dev/google/universal-sentence-encoder/1}{\texttt{universal-sentence-encoder/1}} \\
English & -    & 512 & \href{https://tfhub.dev/google/universal-sentence-encoder/2}{\texttt{universal-sentence-encoder/2}} \\
English & 32B  & 200 &  \\
English & 32B  & 200 &  \\
\bottomrule
\end{tabular}
\end{small}
\end{center}
\vskip -0.1in
\caption{TensorFlow Hub embedding modules for text input.\label{table:hub_modules}}
\end{table*}

\end{document}